%% file: main.tex
\documentclass{article}
\PassOptionsToPackage{numbers,sort&compress}{natbib}
\usepackage[preprint]{neurips_2025}
\usepackage[utf8]{inputenc}
\usepackage[T1]{fontenc}
\usepackage{amsmath,amssymb,booktabs,graphicx,xcolor,nicefrac,microtype}
\usepackage{algorithm}
\usepackage{tikz}
\usetikzlibrary{arrows.meta,positioning,fit,calc}
\definecolor{ours}{HTML}{2A78D6}
\definecolor{contact}{HTML}{EB6834}
\definecolor{muted}{HTML}{52514E}
\definecolor{panel}{HTML}{F3F2EE}
\usepackage{algpseudocode}
\usepackage[colorlinks=true,allcolors=blue]{hyperref}
\usepackage{url}

\title{LevelFirst: Asking About Granularity\\Before Merging in Interactive Image Clustering}
\author{Xin Ma\\Independent Researcher}

\begin{document}
\maketitle

\begin{abstract}
An image collection can support several valid partitions. The same CIFAR-100 images,
for example, can be grouped into 100 fine classes or 20 superclasses. LevelFirst asks
about the user's intended granularity before asking which groups to merge. A
vision-language model proposes a two-level naming hierarchy for graph communities
obtained without a target cluster count. The user decides whether to group or separate
the specific names under each of the three general names that cover the most images.
Each queried branch keeps its own answer, and the majority sets the default for
unqueried branches. Seven pairwise merge queries follow, and a final cleaning step
trains a metric-learning head on pseudo-labels supported by zero-shot name predictions
and neighbor agreement. We evaluate 26 benchmark settings using label-derived
simulated feedback. Before cleaning, LevelFirst achieves a mean
adjusted Rand index (ARI, scaled by 100) of 75.4 with ten answers, compared with 72.9
for ten pairwise merge queries. The gain is concentrated on coarse targets (78.4
vs.\ 66.5), and the scores are close on fine targets (74.5 vs.\ 74.8). After cleaning,
LevelFirst reaches 77.6, compared with 74.5 for calibrated merging with a language-model
judge and the same cleaning. With independent answer flips at probability 0.2,
improvements on coarse targets yield a 4.6-point overall mean gain before cleaning
over the zero-answer partition based on specific names. At a flip probability of 0.1,
LevelFirst misses its prespecified degradation threshold.
\end{abstract}

\section{Introduction}

A clustering is useful only at the granularity its user has in mind. Whether ``calf''
and ``cow'' should share a cluster, or ``fire truck'' and ``garbage truck'', is not a
property of the image features but of the task. One benchmark makes this concrete:
CIFAR-100~\cite{krizhevsky2009cifar} labels its photos with 100 fine classes, and the
same photos also carry 20 superclass labels (CIFAR-100-20). A method that sees only the
photos must return the same partition for both targets, so it is wrong for at least one
of them; TEMI~\cite{adaloglou2023temi} already evaluated both label sets and argued that
the superclasses cannot be inferred from the pixels alone. Images alone do not identify
which valid partition the user intends.

Two lines of work bring it in. Text-assisted pipelines over-segment the data and let a large language model (LLM) decide merges under a textual guideline~\cite{guideline2026agent}, or
cluster captions under a user-written criterion~\cite{kwon2024ictc}. Both set the
granularity through text: the user's wording, which IC$|$TC lets the user revise after
seeing the result, or the LLM's own notion of a kind. Interactive clustering instead asks
the user pairwise ``same group?''
queries~\cite{wagstaff2001copkmeans,basu2004active,vancraenendonck2017cobra}. A pairwise
answer is local. Constraint propagation can carry it beyond the asked
pair~\cite{vancraenendonck2017cobra}, but in the direct contact-merge baseline we compare
with, each ``same'' answer removes at most one group. When an over-segmented start splits ten classes into 30 to 68 groups, as the
vision-language names below do on CIFAR-10 and STL-10, ten answers cannot reach the
target granularity, however well the pairs are chosen.

\begin{figure}[t]
\centering
\begin{tikzpicture}[
  font=\scriptsize, node distance=2.4mm,
  stage/.style={draw=muted, rounded corners=2pt, fill=panel, align=center, text width=23.4mm,
    minimum height=17mm, execute at begin node={\hyphenpenalty=10000}, inner sep=2pt, line width=0.5pt},
  ask/.style={stage, draw=ours, fill=ours!7, line width=0.9pt},
  flow/.style={-{Stealth[length=2mm]}, line width=0.7pt, draw=muted},
  user/.style={-{Stealth[length=2mm]}, line width=0.7pt, draw=ours}]
\node[stage] (base) {{\footnotesize\textbf{Base partition}}\\[2pt]Louvain on a\\neighbor graph;\\no target count};
\node[stage, right=of base] (names) {{\footnotesize\textbf{Name tree}}\\[2pt]a VLM names\\communities;\\CLIP-L assigns\\photos to names};
\node[ask, right=of names] (level) {{\footnotesize\textbf{Level queries}}\\[2pt]the three heaviest\\general names:\\one kind or several?};
\node[stage, right=of level] (merge) {{\footnotesize\textbf{Merge queries}}\\[2pt]seven pairs of\\touching groups:\\same group?};
\node[stage, right=of merge] (clean) {{\footnotesize\textbf{Cleaning}}\\[2pt]confident images\\train an ArcFace\\head; no labels};
\foreach \a/\b in {base/names, names/level, level/merge, merge/clean} \draw[flow] (\a) -- (\b);
\node[above=7mm of $(level.north)!0.5!(merge.north)$, draw=ours, rounded corners=6pt, fill=white,
  inner sep=3pt, font=\footnotesize\bfseries, text=ours] (user) {User: ten answers};
\draw[user] (user.south west) ++(4mm,0) -- (level.north);
\draw[user] (user.south east) ++(-4mm,0) -- (merge.north);
\node[below=3mm of level, font=\scriptsize, align=left, text width=68mm, anchor=north] (tree) {%
  \textbf{truck}: fire truck, garbage truck, dump truck, \ldots\\
  \hspace*{3mm}{\color{ours}$\Rightarrow$ ``one kind'': use \emph{truck}}\\[1pt]
  \textbf{animal}: horse, deer, elk, bird, \ldots\\
  \hspace*{3mm}{\color{contact}$\Rightarrow$ ``several kinds'': keep the specific names}};
\draw[muted, densely dotted, line width=0.5pt] (level.south) -- (tree.north);
\end{tikzpicture}
\caption{The pipeline. The first three of the user's ten answers set the granularity of
the name vocabulary; the remaining seven merge name groups; cleaning then removes
over-split groups without labels. The example names are those the VLM produced on STL-10
(Figure~\ref{fig:example}).}\label{fig:overview}
\end{figure}
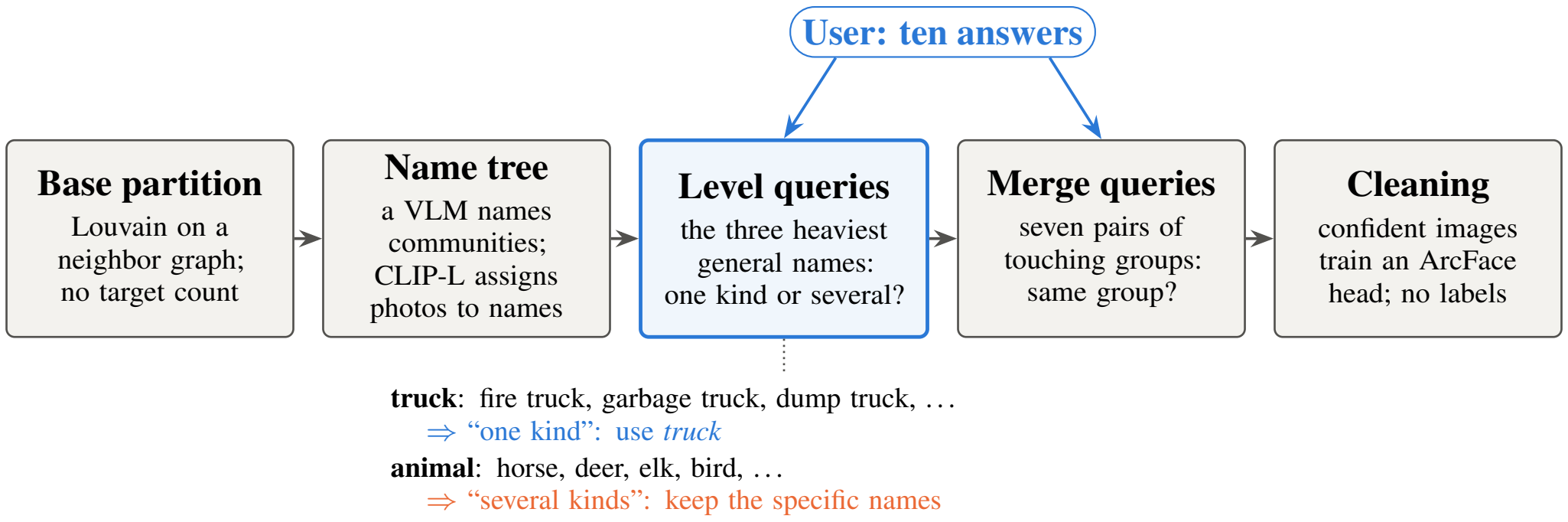

We ask about granularity directly. A vision-language model (VLM) names each community of
a count-free graph clustering with a specific noun (``fire truck'') and a general noun
(``truck''), and these names form a two-level tree. The user is shown the three general
names that cover the most images, each with its specific names, and says whether they
are one kind or several. One such answer acts on every group under that general name,
and the majority of the three answers sets the level of the rest of the vocabulary, so
three answers can change the granularity of the whole partition. The remaining answers
go to pairwise merges, and a label-free cleaning step removes over-split groups
(Figure~\ref{fig:overview}). We call this method \textbf{LevelFirst}.

\paragraph{Contributions.}
\begin{enumerate}
\item \textbf{Level queries.} We introduce queries about a VLM name tree that
calibrate granularity before any merge. Three of them raise name-group ARI by 17 to 30
points on CIFAR-10 and STL-10 and by 13.5 on CIFAR-100-20, and leave CIFAR-100 and Tiny
ImageNet essentially unchanged (Section~\ref{sec:levels}).
\item \textbf{An evaluation under declared criteria.} On 26 target sets from five
datasets, with ten answers and before cleaning, LevelFirst reaches 75.4 mean ARI vs.\
72.9 for ten contact queries (78.4 vs.\ 66.5 on coarse and 74.5 vs.\ 74.8 on fine
targets). After cleaning it reaches 77.6 vs.\ 74.5 for a calibrated-merge pipeline with an
LLM judge, and on fine targets it is 0.3 lower (77.1 vs.\ 77.4). Providing the true class
count did not close the gap for the three geometric references we tested (at most 61.4).
Six of the fine sets are class splits first evaluated in the final comparison
(Section~\ref{sec:results}).
\item \textbf{Robustness and negative results.} We flip each simulated answer
independently with probability 0.1 to 0.3, and test two answer-checking rules. We also report the
declared variants that failed, including calibrated merging after level queries and
validated automatic merges (Sections~\ref{sec:noise} and~\ref{sec:negative}).
\end{enumerate}

The method itself, including the three-query level rule, was developed on eight
development sets. Every evaluation reported after that was declared with its pass
criteria before it ran, and we report failed criteria next to passed ones. The user is simulated from labels, and the VLM and CLIP-L
add semantic knowledge that feature-only methods lack; all query strategies and
name-based baselines share these resources (Section~\ref{sec:setting}). Code, declarations and per-set results will be
released.

\section{Related work}\label{sec:related}

\paragraph{Image clustering with a known target.}
Deep image clustering methods learn a partition into a given number of
clusters~\cite{vangansbeke2020scan,niu2022spice,ma2022lnscc}. With pretrained features,
TEMI~\cite{adaloglou2023temi} and TURTLE~\cite{gadetsky2024turtle} reach high accuracy on
standard benchmarks, and TAC~\cite{li2024tac} and SIC~\cite{cai2023sic} add CLIP text
embeddings of nouns~\cite{radford2021clip}. All of them take the number of clusters as
input and evaluate against one label set. Count estimators such as the gap
statistic~\cite{tibshirani2001gap} choose a count from the data alone, so they return the
same answer for CIFAR-100 and CIFAR-100-20. We do not assume the count and ask the user for the granularity instead. We compare with simple known-count references on the same features (Section~\ref{sec:results}); published known-count numbers use other encoders and protocols and are not directly comparable.

\paragraph{Over-cluster, then merge with an LLM.}
A hybrid LLM agent clusters guideline-focused caption embeddings with
HDBSCAN~\cite{campello2013hdbscan} into small consistent clusters, then lets an LLM judge
merges along a minimum spanning tree~\cite{guideline2026agent}. It needs no training and
no user answer during clustering; granularity follows from the guideline and the LLM's
judgments. We compare with an adapted reproduction of its minimum-spanning-tree merge
step on our name groups; we do not reproduce its guideline-conditioned caption
embeddings.

\paragraph{Granularity and user criteria.}
ClusterLLM lets an LLM answer triplet and pair queries to tune text embeddings and to
choose the granularity of text clusters~\cite{zhang2023clusterllm}. IC$|$TC clusters
images by captions under a user's text criterion~\cite{kwon2024ictc}. Multiple-clustering
methods bias embeddings toward a user keyword~\cite{multisub2024} or propose criteria
automatically~\cite{xcluster2024}. Closest to our setting, IPC~\cite{liu2026ipc}
clusters images under a user's natural-language criterion and then refines the groups
with the user's feedback on selected sample pairs. Our level queries differ in two ways:
unlike ClusterLLM, they ask the user rather than an LLM, and unlike IC$|$TC and IPC, they
ask the user to choose the coarse or fine names of branches in a name tree generated from
the images, rather than to state a criterion in advance.

\paragraph{Interactive and constrained clustering.}
Must-link and cannot-link constraints~\cite{wagstaff2001copkmeans} and their active
selection~\cite{basu2004active} are classic. Split and merge requests on whole clusters
have query-complexity guarantees~\cite{balcan2008split}. COBRA over-clusters into
super-instances and merges them with pairwise queries, propagating the
constraints~\cite{vancraenendonck2017cobra}; our merge queries follow its
over-cluster-and-merge design without propagation and form the main interactive
baseline. A level
query is closer to a merge request on many clusters at once: one answer decides every
group under a general name.

\paragraph{Self-labeling and noisy labels.}
Our cleaning step trains on reliable pseudo-labels, as in
self-labeling~\cite{vangansbeke2020scan,niu2022spice}, and selects them by neighbor
agreement, as deep k-NN does for noisy labels~\cite{bahri2020knn}; DivideMix instead
separates samples by a mixture model of their losses~\cite{li2020dividemix}. We claim no
novelty for it and evaluate it only as part of the pipeline.

\section{Setting}\label{sec:setting}

\paragraph{Task.}
A collection of $n$ images is to be partitioned into the user's classes. Their number is
unknown. The method may ask the user $B=10$ queries. We report the adjusted Rand
index~\cite{hubert1985ari} (ARI) as the primary metric, and accuracy under Hungarian
matching (ACC) and NMI as secondary metrics.

\paragraph{Simulated user.}
Answers come from an unflipped label-derived simulator. A \emph{merge query} shows two
groups, and the answer is ``same'' exactly when their majority classes agree. A
\emph{level query} shows a general name with its $m$ specific names, and the answer is
``several kinds'' when the occupied specific-name groups have more than $m/2$ distinct
majority classes; for example, four names whose groups have only two distinct majority
classes receive ``one kind''. Section~\ref{sec:noise} flips these answers at random.

\paragraph{Use of labels.}
Benchmark labels define the balanced and long-tail sets, generate the simulated answers
and serve evaluation. They also informed the choice of the three-query level rule on the
development sets (below), give the true class count to the known-count references, and
supply class names to the oracle vocabulary diagnostic of Section~\ref{sec:levels}, which
is labeled as such. Cleaning trains on pseudo-labels from the current partition, and the
name-based methods generate their vocabulary from the images without a list of benchmark
class names.

\paragraph{Resources.}
The base features concatenate unit-normalized DINOv2 ViT-S/14~\cite{oquab2023dinov2} and
CLIP ViT-B/32 image embeddings. The \emph{added semantic resources} are
Qwen2.5-VL 7B~\cite{bai2025qwen25vl}, which proposes names; the CLIP ViT-L/14 (CLIP-L) image and text towers, which assign images to names; and gpt-oss 20B~\cite{openai2025gptoss}, used
only by the baselines marked with $\dagger$. No benchmark class name enters any prompt,
or any vocabulary except that of the oracle diagnostic. Feature-only baselines such as
Louvain do not use these resources, so we report them as references, not as competitors
at equal resources.

\paragraph{Datasets and sets.}
Each dataset gives a \emph{balanced} set and a \emph{long-tail} set with exponentially
decreasing class sizes (imbalance ratio 100). On CIFAR-100~\cite{krizhevsky2009cifar},
each of seven class splits (0--6) targets a seeded half of the classes, giving 50
classes and 25{,}000 (balanced) or 5{,}499 (long-tail) images. Tiny
ImageNet~\cite{le2015tiny} has three class splits (0--2) of 100 classes with 50{,}000 or
10{,}847 images. CIFAR-10~\cite{krizhevsky2009cifar} uses its 50{,}000 training images
and STL-10~\cite{coates2011stl} its 13{,}000 labeled images, both with all ten classes.
CIFAR-100-20 uses all 50{,}000 CIFAR-100 training images with their 20 superclass
labels. This gives 26 sets: 20 \emph{fine} sets (CIFAR-100 and Tiny ImageNet) and 6
\emph{coarse} sets (CIFAR-10, STL-10 and CIFAR-100-20). Unlike the example in the
introduction, the fine CIFAR-100 sets use half of the classes and the coarse CIFAR-100-20
sets all of them, and long-tail sets are sampled separately, so fine and coarse targets
are evaluated on separately constructed sets rather than on one collection.

\paragraph{Roles of the sets.}
The level rule (three queries, unweighted majority) was chosen on eight
\emph{development} sets: CIFAR-100 split 0, Tiny ImageNet split 0, CIFAR-10 and STL-10,
each balanced and long-tail. It was then evaluated under declared criteria on twelve \emph{held-out} sets: CIFAR-100
splits 1--4, Tiny ImageNet split 1 and CIFAR-100-20. Six \emph{fresh} sets (CIFAR-100
splits 5--6, Tiny ImageNet split 2) are class subsets first evaluated afterwards, in a
final declared comparison of pipelines. Because splits are class subsets of the same
training images, fresh and held-out sets can share images with development sets. All six
fresh sets are fine, and four of the six coarse sets (CIFAR-10 and STL-10) are development
sets, so only CIFAR-100-20 gives held-out coarse evidence. The merge queries and the
cleaning loop were developed earlier on CIFAR-100 splits 0--2, before the level rule
existed.

\section{Method}\label{sec:method}

Each image $i$ has a base feature $x_i$ (DINOv2 ViT-S/14 and CLIP ViT-B/32 embeddings, concatenated) and a CLIP-L
image embedding $z_i$; both are unit-normalized. $N(i)$ denotes the ten nearest neighbors of
$x_i$ by cosine similarity. LevelFirst spends the budget of $B=10$ answers as three level
queries followed by seven merge queries (Figure~\ref{fig:overview}).

\subsection{Base partition and name tree}
\paragraph{Base partition.}
Louvain~\cite{blondel2008louvain} runs on the shared-neighbor graph of $N$, at resolution 1
and without a target count. It gives 22 to 74 communities.

\paragraph{Names.}
Each community is split into three KMeans pieces (one piece below 16 images), and the eight images nearest each piece's centroid form a grid. The VLM sees the grids of one community in a single call
and lists at most four kinds of object, each with a \emph{specific} and a \emph{general}
common noun of one to three words. After normalization, every specific name becomes a
leaf under its general name; a specific name listed under two general names is a leaf of
both, and a vocabulary keeps each name once. The result is a two-level name tree $T$ that maps each
general name $g$ to its specific names $S_g$.

\paragraph{Assignment to names.}
A \emph{vocabulary} $V$ is a set of names drawn from $T$. Each name $v \in V$ is embedded
as $t_v$ with the CLIP-L text tower, averaged over seven prompt templates. Image $i$ in
community $c$ takes the name
\[
a_i = \arg\max_{v \in V} \; 100\, z_i^\top t_v + \log\big(\pi_c(v) + 10^{-3}\big),
\]
where $\pi_c(v)$ is the share of $c$'s images whose plain zero-shot name $\arg\max_v z^\top
t_v$ is $v$. The prior keeps a name that is rare in a community from capturing its images.
Images with the same name form a \emph{name group}. With the specific vocabulary
$\bigcup_g S_g$ there are 30 to 201 name groups: mostly pure, but over-split through
synonyms, subtypes and scene nouns.

\subsection{Level queries}\label{sec:levelmethod}
\begin{figure}[t]
\centering
\includegraphics[width=\linewidth]{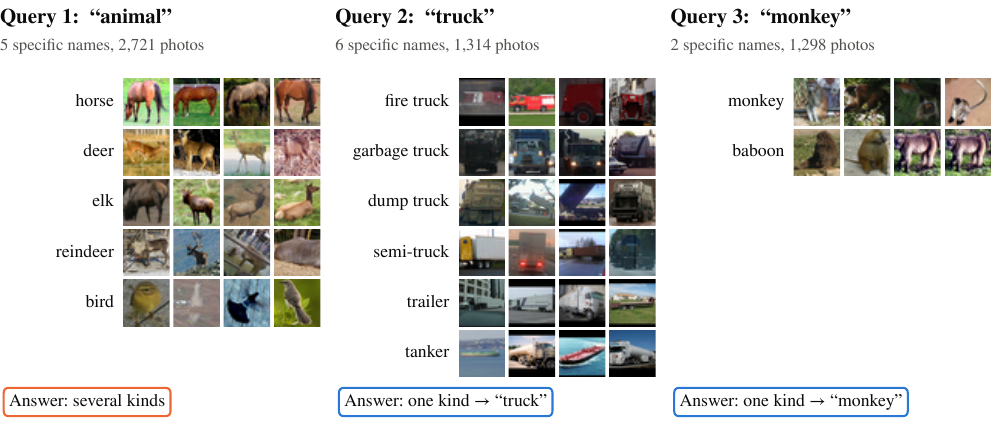}
\caption{The three level queries on STL-10 (balanced). Each row shows the four images with the highest CLIP-L similarity among those assigned to that specific name; the selection uses no label. The answers are those of the simulated user, computed from labels
(Section~\ref{sec:setting}). The majority (``one kind'') then sets the general level for
every other general name; this raised name-group ARI on this set from 46.8 to 73.1.
Photos from STL-10~\cite{coates2011stl}, which draws its images from
ImageNet~\cite{deng2009imagenet}.}\label{fig:example}
\end{figure}

Algorithm~\ref{alg:levels} spends the first three answers.
\begin{itemize}
\item A \emph{candidate} is a general name with at least two specific names. Its weight
$w_g$ is the number of images assigned to its specific names, so the heaviest candidates
cover the most images.
\item The user is shown each of the three heaviest candidates with its specific names and
answers ``one kind'' or ``several kinds''. An interface would show example images for each specific name (Figure~\ref{fig:example}); the simulated user answers from labels.
\item Each asked general name keeps its own answer, and every other general name takes
the majority of the three.
\end{itemize}
``One kind'' replaces the specific names $S_g$ by $g$ in the vocabulary, and all images
are reassigned with the rule above. A single answer thus changes every group under $g$,
and the majority extends it to the rest of the tree. The majority is unweighted: in
development, a catch-all general name (``animal'' on CIFAR-10 long tail) carried most of
the image weight and flipped an image-weighted vote.

\begin{algorithm}[t]
\caption{Level queries}\label{alg:levels}
\begin{algorithmic}[1]
\Require name tree $T$ (general $g$ $\to$ specifics $S_g$), assignment $a$ of images to specific names
\State $C \gets \{g : |S_g| \ge 2\}$, \quad $w_g \gets |\{i : a_i \in S_g\}|$
\State $Q \gets$ the three $g \in C$ with largest $w_g$ (ties by name)
\For{$g \in Q$} \State ask: ``Are $S_g$ one kind ($g$) or several kinds?'' \quad $\ell_g \gets$ answer \EndFor
\State $d \gets$ majority of $\{\ell_g : g \in Q\}$
\State $\ell_g \gets d$ for every $g \notin Q$
\State vocabulary $\gets \bigcup_g (S_g \text{ if } \ell_g = \text{several, else } \{g\})$; reassign images with CLIP-L
\end{algorithmic}
\end{algorithm}

\subsection{Merge queries}\label{sec:mergemethod}
The remaining seven answers go to pairwise \emph{contact queries} on the name groups.
Two groups $A$ and $B$ touch when some image in one has a neighbor in the other; let
$e(A,B)$ count the directed neighbor edges between them in both directions. Pairs are
asked in decreasing order of the priority $\min(|A|,|B|)\cdot\kappa(A,B)$, with ties broken
by group index, where $\kappa(A,B) = e(A,B) / (20 \min(|A|,|B|))$ is a normalized
crossing-edge count. It can exceed 1, because edges from the larger group into the smaller
one also count. The priority equals $e(A,B)/20$, so large, strongly touching groups are
asked first. ``Same'' merges the two groups and
recomputes the candidates; ``different'' is remembered for that pair.

\subsection{Label-free cleaning}\label{sec:cleanmethod}
Cleaning starts from the ten-answer partition $P$ and never reads a label.

\paragraph{Confident images.}
Each specific name group maps to the majority $P$-cluster of its images. Image $i$ is
\emph{confident} when both of these hold:
\begin{itemize}
\item its plain zero-shot specific name maps to $P(i)$;
\item the majority of $P$ over $N(i)$ is $P(i)$.
\end{itemize}

\paragraph{Training.}
An ArcFace head~\cite{deng2019arcface} on $x_i$ is trained on the confident images with
their $P$-clusters as classes, and every image then takes the head's argmax. The head is
an MLP of size 896--512--256 with scale 30 and margin 0.3, trained for 30 epochs with
batch size 256, learning rate $10^{-3}$ and seed 0.

\paragraph{Iteration.}
The round runs four times, each on the previous round's partition. From round 2, $N(i)$
is recomputed in the head's embedding. The output is the first partition from round 2 on
that agrees with its predecessor at ARI $\ge 0.98$, or the fourth if none does.

\paragraph{Effect on clusters.}
A cluster with no confident image receives no class, so cleaning can remove over-split
groups. It cannot create a new cluster and so cannot split a merged one, although the
per-image prediction can move images between surviving clusters.

\paragraph{Cost.}
Naming takes one VLM call per community, and the name assignment one CLIP-L text encoding
per vocabulary change. Level and merge queries call no model. Each cleaning round trains one small head on the base features.

\section{Experiments}\label{sec:results}

Every method runs once on the 26 sets of Section~\ref{sec:setting} with the simulated
user; only the noise study (Section~\ref{sec:noise}) repeats runs over five seeds. We
report ARI in \%; Table~\ref{tab:secondary} adds ACC and NMI for the final
pipeline.

\subsection{Baselines}
All baselines start from the same base partition and, where they use names, from the same
name tree.
\begin{itemize}
\item \textbf{Contact queries.} Ten merge queries on the specific name groups.
\item \textbf{Calibrated merge queries$^\dagger$.} A logistic calibrator chooses the
merge queries and links pairs automatically above probability 0.9, once a
leave-one-out check passes. It is fitted on the answers, on cheap pair features, and on
three gpt-oss judge answers for the 240 highest-prior name pairs. Followed by cleaning,
it was the strongest pipeline before LevelFirst.
\item \textbf{LLM merge$^\dagger$.} An adapted reproduction of the merge step
of~\cite{guideline2026agent}, run on our name groups:
gpt-oss judges merges of the name groups along a Ward minimum spanning tree, from VLM
captions of the five images nearest each group centroid. It works under a generic
guideline, ``Group the photos by the kind of object they mainly show.'' Its departures
from the original (judge model, caption handling, noise points) are documented in the
repository.
\item \textbf{Name groups + cleaning.} The same cleaning loop without answers.
\item \textbf{Known-count references.} Methods that receive the true class count $K$
instead of answers: KMeans on the base features, Louvain on the same shared-neighbor
graph with its resolution bisected to come as close as possible to $K$ communities, and
KMeans on CLIP-L image
embeddings. KMeans uses ten initializations and seed 0.
\end{itemize}

\subsection{Main results}
Table~\ref{tab:main} reports mean ARI over the 26 sets, by dataset and regime.
LevelFirst gains on coarse targets, where the specific names are far too fine. On fine
targets, where the level queries mostly keep them, it ends 0.3 ARI below the best
alternative both before and after cleaning.

\begin{table}[t]
\centering\small
\caption{Mean ARI (\%) by dataset group, with the number of sets in parentheses. Fine
targets are CIFAR-100 (C-100) and Tiny ImageNet (Tiny IN); coarse targets are CIFAR-10,
STL-10 and CIFAR-100-20 (C-100-20). Bold marks the best value in each column within each
block of the same answer budget and cleaning stage; the two reference blocks are not
ranked.
$K$: the true class count is given instead of answers. $^\dagger$Uses the gpt-oss judge in addition to the shared resources.}\label{tab:main}
\smallskip
\setlength{\tabcolsep}{3.5pt}
\input{generated/main_table.tex}
\end{table}

\paragraph{Coarse targets.}
Level queries change coarse targets the most. With ten answers, LevelFirst reaches
78.4 ARI before cleaning, compared with 66.5 for ten contact queries and 62.6 for
calibrated merging. After cleaning it reaches 79.4 vs.\ 64.7 for the calibrated-merge
pipeline, and 50.8 vs.\ 33.2 on CIFAR-100-20. It is ahead on all six coarse sets, by up to
26.5 ARI. Without level queries, the VLM's specific names over-split ten classes into 30 to 68
groups. Each merge answer removes at most one group, so neither seven nor ten merge
queries can undo that.

\paragraph{Fine targets.}
On fine targets the level queries mostly keep the specific names (16 of 20 sets).
Before cleaning, LevelFirst is 0.3 ARI below ten contact queries (74.5 vs.\ 74.8) and
lower on 15 of 20 sets, by at most 1.0 ARI, because three answers go to level queries
that change little here. After cleaning it is 0.3 below the calibrated-merge pipeline
(77.1 vs.\ 77.4), lower on 13 sets and higher on 6, with per-set differences between
$-2.9$ and $+2.5$ ARI. LevelFirst therefore does not improve fine targets; its benefit is that it does not need
to know in advance which kind of target it faces.

\begin{figure}[t]
\centering
\includegraphics[width=\linewidth]{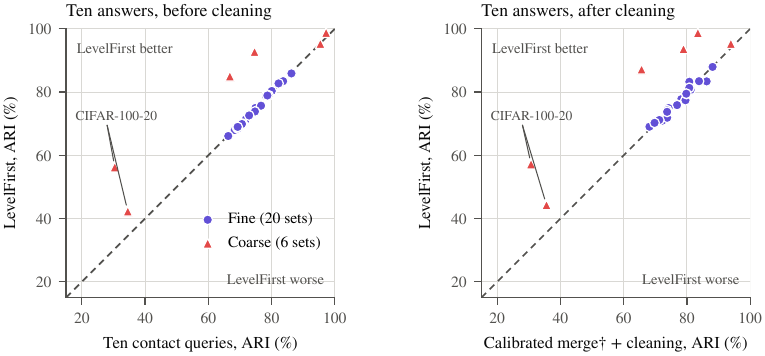}
\caption{Per-set ARI at ten answers: LevelFirst vs.\ ten contact queries, both before cleaning (left), and vs.\ the calibrated-merge pipeline, both after cleaning (right). Points above the diagonal favor
LevelFirst. Fine sets lie close to the diagonal; most coarse sets lie well above it.}\label{fig:paired}
\end{figure}

\begin{figure}[t]
\centering
\includegraphics[width=\linewidth]{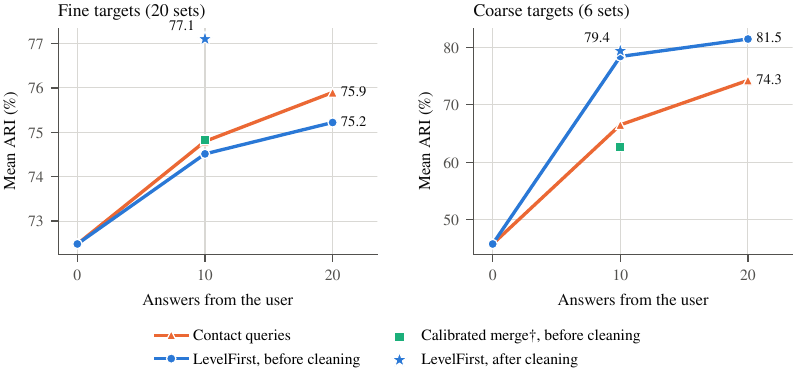}
\caption{Mean ARI vs.\ the number of answers on fine (left) and coarse (right) targets. The panels use different vertical ranges, and lines connect the measurements at 0, 10 and 20 answers. With zero answers both query strategies start from the specific name groups. The square is calibrated merging before cleaning and the star LevelFirst after cleaning, both with ten answers; all other points are before cleaning.}\label{fig:budget}
\end{figure}

\paragraph{Overall.}
Over all 26 sets, LevelFirst reaches 75.4 ARI before cleaning, compared with 72.9 for ten
contact queries. After cleaning it reaches 77.6, compared with 74.5 for the
calibrated-merge pipeline and 69.9 for cleaning without answers, and it needs no LLM
judge.
Figure~\ref{fig:paired} shows every set, and Figure~\ref{fig:budget} the effect of the
answer budget: on coarse targets twenty contact queries stay below ten answers that
start with level queries, while on fine targets the two strategies stay within one ARI
point.

\paragraph{Known class count.}
Providing the true class count did not close the gap for the three geometric references
we tested (Table~\ref{tab:main}). The best of them, count-matched Louvain, reaches 61.4
ARI over all sets (58.7 fine, 70.6 coarse), compared with 77.6 for LevelFirst with ten
answers, and none beats LevelFirst on a fine set. The comparison changes more than the
count: the references use no names, answers or cleaning. KMeans on CLIP-L, the image
space that LevelFirst uses for names, is the weakest (48.9), so the count alone does not
recover what the names provide. Count-matched Louvain reached $K$ exactly on 25 of 26
sets and 49 communities for 50 classes on CIFAR-100 split 4 long tail. The exception is
STL-10 balanced: all three references beat LevelFirst there, KMeans on the base features
reaching 97.7 vs.\ 87.2, and LevelFirst ends with 19 clusters for 10 classes. These are simple references on the same
features; tuned known-count methods with larger encoders, such as
TURTLE~\cite{gadetsky2024turtle}, are stronger and are not compared here.

\paragraph{The LLM merge baseline.}
Its adapted merge step leaves the mean ARI of our name groups unchanged (66.3 for both). Its caption judge
rejects most merges, including most same-class pairs.

\paragraph{ACC, NMI and cluster counts.}
Table~\ref{tab:secondary} reports them for LevelFirst.

\begin{table}[t]
\centering\small
\caption{LevelFirst after cleaning (ten answers): ARI, ACC and NMI (\%), mean final
cluster count and class count, and in round 1 the share of confident images (Kept) and their purity, the share whose class is the majority class among the confident images of their cluster.}\label{tab:secondary}
\input{generated/secondary.tex}
\end{table}

\subsection{Declared evaluations}
Two declared evaluations bear on the method.

\paragraph{Held-out evaluation (twelve sets).}
Every primary criterion passed (Appendix~\ref{app:criteria}). Before cleaning, LevelFirst
reached 70.6 ARI vs.\ 68.1 for ten contact queries (bar 70.1), and on the ten fine sets
it stayed within 0.5 of them (74.8 vs.\ 75.2). After cleaning it reached 73.2 vs.\ 71.1
for the calibrated-merge pipeline (bar 73.1). On CIFAR-100-20 the level queries chose
the general level on both sets, and LevelFirst beat contact queries by 16.8 ARI before
cleaning.

\paragraph{Fresh comparison of pipelines.}
On the held-out fine sets the calibrated-merge pipeline had ended 1.0 ARI above
LevelFirst, so we declared a combined variant, level queries followed by calibrated merge
queries, and tested it on the six fresh fine sets and the six coarse sets. It failed
every criterion. On the fresh fine sets the three pipelines ended at 76.6 (combined), 77.6
(LevelFirst) and 77.0 (calibrated merge), so the earlier fine-set lead of calibrated
merging did not replicate. On the coarse sets the combined variant ended 1.6 below
LevelFirst, and at ten answers the calibrator made no automatic merge on any of the 26
sets. LevelFirst therefore keeps plain contact queries after the level queries.

\subsection{What the level queries decide}\label{sec:levels}

\begin{table}[t]
\centering\small
\caption{Level decisions and name-group ARI (\%) with the specific names and after the three level queries, before any merge query. ``Specific level'' counts the sets whose majority answer kept the specific names.}\label{tab:levels}
\input{generated/levels.tex}
\end{table}

\begin{figure}[t]
\centering
\includegraphics[width=\linewidth]{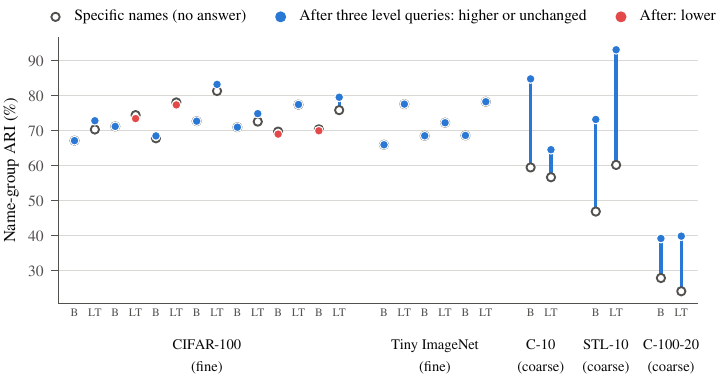}
\caption{Name-group ARI of every set before (hollow) and after (filled) the three level
queries, before any merge query. B: balanced, LT: long tail.}\label{fig:levels}
\end{figure}

Table~\ref{tab:levels} and Figure~\ref{fig:levels} isolate the level step.

On every coarse set the majority answer chose the general names, which raised name-group
ARI by 16.6 on CIFAR-10, 29.6 on STL-10 and 13.5 on CIFAR-100-20. Every Tiny ImageNet set
kept the specific names. On CIFAR-100, four of fourteen sets chose the general level, yet
ARI rose slightly (+0.6): the general names the VLM uses on CIFAR-100 are mostly
fine-grained already, so a wrong-looking ``one kind'' answer costs little there.

\paragraph{An oracle reference.}
We also evaluated an aggregate oracle: an image-weighted vote over the verdicts on
\emph{every} general name, counted as one answer and followed by nine contact queries.
On the held-out sets it reached 70.5, compared with 70.6 for the three-query rule with
seven contact queries. The two differ in both the level decision and the number of merge
queries, so the scores do not isolate the aggregation; three answers a user can give
reach the same overall level as this oracle.

\paragraph{CIFAR-100-20 remains hard.}
Superclasses such as ``large omnivores'' join visually distinct images. Even an oracle
superclass vocabulary reaches only 45.9 and 69.6 ARI on the two sets.

\subsection{Noisy users}\label{sec:noise}

\begin{table}[t]
\centering\small
\caption{ARI (\%) with ten answers, before cleaning, when every answer is flipped with probability $p$ (shared random draws across rates; means over 26 sets and five seeds). Bottom: LevelFirst after cleaning on the twelve sets of the fresh comparison (six fresh fine and six coarse sets), with one noise seed (seed 0).}\label{tab:noise}
\input{generated/noise.tex}
\end{table}

\begin{figure}[t]
\centering
\includegraphics[width=0.62\linewidth]{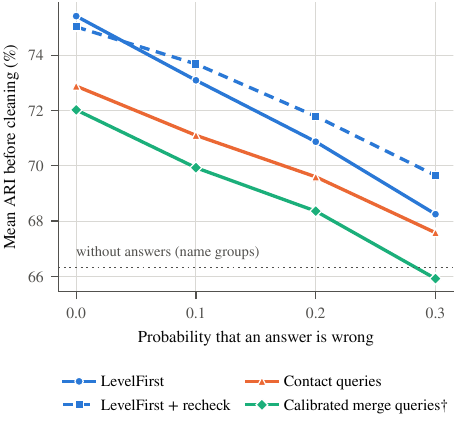}
\caption{Mean ARI with ten answers, before cleaning, over 26 sets and five seeds, vs.\ the
probability of a wrong answer. The dashed line adds the exploratory recheck rule
(Section~\ref{sec:noise}); the dotted line is the specific name groups without answers.}\label{fig:noise}
\end{figure}

To assess sensitivity to response errors, we independently flipped each simulated answer,
level and merge alike, with probability $p$. Each strategy draws one uniform number per
query position, shared across rates; because later queries depend on earlier answers, the
queries asked can differ between rates. Table~\ref{tab:noise} and
Figure~\ref{fig:noise} show the result.

\paragraph{Declared criteria.}
Two of the three primary criteria passed. At $p=0.2$, LevelFirst stays 4.6 ARI above the
name groups without answers before cleaning (70.9 vs.\ 66.3). This mean comes from the
coarse targets: the gain is 22.4 ARI on the six coarse sets but $-0.8$ on the 20 fine
sets, 15 of which fall below their own zero-answer name groups. After cleaning, on the
twelve sets of the fresh comparison, it stays 10.9 ARI above cleaning without answers.
The graceful-degradation criterion failed narrowly: at $p=0.1$ the score before cleaning
fell to 73.1, below its bar of 73.4.
LevelFirst remained the best of the three query strategies at every rate.

\paragraph{The cost of wrong answers.}
Each 10\% of wrong answers costs about 2 ARI, and every strategy pays this cost. The
level vote is associated with a specific risk: at $p=0.1$, runs whose level default
flipped lost 8.7 ARI on average (16 on coarse sets), against 1.8 for the other runs.
Flipped defaults occurred in 7\% of runs but carried a quarter of the loss. This post hoc
grouping of observed runs is an association, not a measured cost, since flipped runs also
differ in their later merge queries.

\paragraph{Answer checking.}
After inspecting these results on the same 26 sets, we devised two checks for split level
answers, both paid from the ten answers, and declared their criteria before running them;
this follow-up is therefore exploratory.
\emph{Escalate} asks two more general names. It failed, because the fourth and fifth
heaviest names are weaker evidence and reversed correct defaults on fresh fine sets.
\emph{Recheck} asks the three queries again and drops inconsistent ones. It passed its
primary criteria (73.7 vs.\ 73.1 at $p=0.1$), but only under the optimistic assumption
that a repeated query gets an independent error, and it missed its secondary goal of
halving flipped defaults: they rose from 6.9\% to 9.2\% of runs. LevelFirst keeps the plain majority.

\subsection{Negative results}\label{sec:negative}

\paragraph{Calibrated merging.}
With ten answers, a calibrator over judge features matched contact queries on fine
targets (74.8 for both) but fell behind on coarse ones (62.6 vs.\ 66.5). On CIFAR-100 split 0, only 3.7\% and 6.4\% of candidate pairs join groups of the same
class, too few positives to learn from ten answers. Validated automatic merges were
unstable: in the combined variant on Tiny ImageNet split 1 balanced, after twenty answers,
the calibrator proposed 123 automatic links, of which 39.8\% agreed with the simulator's
majority-class rule; with the manual merges, the partition fell from 193 to 86 clusters
and to 24.6 ARI. The same variant reached 70.2 vs.\ 58.8 for level and contact queries
on CIFAR-100-20 long tail at twenty answers; its final snapshot records no automatic
link.

\paragraph{LLM judges of name pairs.}
In earlier experiments on CIFAR-100, a text judge asked whether two names are the same
kind was stricter than the user. Given the user's earlier answers as examples, it became
too loose, and its automatic merges were only 40\% to 50\% correct.

\paragraph{Limits of cleaning.}
Iterating with a fixed neighbor graph gains nothing over one round; only recomputing
neighbors in the learned embedding helps, by about 1 ARI. Without answers, cleaning adds
about 3.6 ARI to the name groups but cannot fix their granularity: on coarse sets it ends
at 49.5, compared with 79.4 for LevelFirst.

\section{Limitations}\label{sec:limitations}

\paragraph{Evaluation.}
\begin{itemize}
\item \textbf{Simulated users.} Answers come from labels, and the noise model flips
answers symmetrically and independently. Real users probably err most on confusable
names, and a wrong level majority is the costliest error (Section~\ref{sec:noise}). A
study with people is the decisive next test.
\item \textbf{Single runs.} Each condition ran once, except in the noise study. CPU
training alone moves a cleaning result by up to about 1 ARI, so per-set differences below
that are not meaningful.
\item \textbf{Datasets.} All images are 32 to 96 pixels, and the six coarse sets come
from three datasets, four of them development sets. We have not tested ImageNet-scale or fine-grained collections such
as birds or cars.
\end{itemize}

\paragraph{Method.}
\begin{itemize}
\item \textbf{Added resources.} The VLM names and CLIP-L carry semantic knowledge that
feature-only methods lack. Our comparisons hold these resources fixed across methods; they
are not comparisons with fully unsupervised clustering.
\item \textbf{Dependence on the names.} The level queries can only choose between the
VLM's specific and general names. On CIFAR-100-20 even an oracle superclass vocabulary
reaches only 45.9 and 69.6 ARI, and our final partitions keep 82 and 50 clusters for 20
classes.
\item \textbf{Two levels, merges only.} Targets between the two levels, or orthogonal to
object identity (for example color or scene), are outside the method. Merge queries only join groups, and cleaning can move images between existing clusters
but never creates a new cluster, so no step splits a merged class into a new cluster; on all ten fine long-tail sets the final
count (31 to 69) is below the class count (50 or 100).
\item \textbf{Cost on fine targets.} When the specific names already fit, the three level
answers change little, and LevelFirst before cleaning is lower than ten merge queries on 15
of 20 fine sets, by at most 1.0 ARI.
\end{itemize}

\paragraph{Comparisons.}
The LLM merge baseline is an adapted reproduction. We have not compared with
IC$|$TC~\cite{kwon2024ictc}, IPC~\cite{liu2026ipc} or active constraint
selection~\cite{basu2004active} at an equal budget, and our known-count references are simple ones on the same features, not tuned known-count methods such as TURTLE (Section~\ref{sec:related}).

\section{Conclusion}
When the same photos can be clustered at several granularities, the first few user
answers are best spent on granularity. LevelFirst asks three queries about a VLM's name
tree before any merge. On all six coarse sets the majority chose the general level, and
on fine sets the level step changed ARI little. With seven merge queries and label-free
cleaning, the pipeline improves coarse targets by 14.7 ARI over the calibrated-merge
pipeline, ends 0.3 ARI below it on fine targets, and needs no LLM judge. With each answer
flipped independently with probability 0.2, its mean gain over no answers remains, but
only on coarse targets. A study with
real users, and name trees with more than two levels, are the natural next steps.

\paragraph{Code availability.}
The code at a fixed revision, the dated experiment declarations with their pass
criteria, the compact per-set results and the table and figure scripts will be released.
Every table and result figure can be regenerated from the compact results alone;
replaying the experiments requires the model caches and run artifacts.

\bibliographystyle{plainnat}
\bibliography{references}

\clearpage
\appendix
\section{Declared criteria}\label{app:criteria}
Each evaluation fixed its sets, methods, criteria and reading rules in a declaration before
its first run. Table~\ref{tab:criteria} lists every criterion with its value and bar. A primary criterion decides how the result is read; a secondary one is reported but does not decide.

\begin{table}[!h]
\centering\footnotesize
\caption{All declared criteria of the four evaluations behind LevelFirst. Two bars mean
the value must clear both. Rows with ``--'' compare methods rather than one value with a bar. Values are mean ARI (\%) over the stated sets, except the flipped-default rows (\% of runs).}\label{tab:criteria}
\input{generated/criteria.tex}
\end{table}

\clearpage
\section{Per-set results}\label{app:perset}
Table~\ref{tab:perset} lists every set. ``Role'' is the set's role in the declared
evaluations: development (dev.), held-out or fresh. ``Level'' is the default level that
the three level answers chose: the specific (spec.) or the general (gen.) names. Bold
marks the better of the two ten-answer methods at the same stage, before or after cleaning.

\begin{table}[!h]
\centering\footnotesize
\setlength{\tabcolsep}{3.2pt}
\caption{ARI (\%) per set. B: balanced; LT: long tail. Louv.: Louvain base; Names:
specific name groups; LLM: LLM merge baseline; Cont.: ten contact queries; LF in.: LevelFirst
before cleaning; Cal.+cl.: calibrated merge and cleaning; LF: LevelFirst after cleaning. Level: the default level chosen by the three level queries, specific (spec.) or general (gen.). Bold marks the better of the two methods with ten answers before cleaning (Cont., LF in.) and after cleaning (Cal.+cl., LF). $^\dagger$Uses the gpt-oss judge.}\label{tab:perset}
\input{generated/per_set.tex}
\end{table}

\clearpage
\section{Prompts and templates}\label{app:prompts}
The prompts below are copied by script from the source code. Angle brackets mark the
parts filled in at run time. No prompt contains a benchmark class name.

\input{generated/prompts.tex}

\end{document}

%% file: generated/main_table.tex
\begin{tabular}{lcrrrrrrr}
\toprule
 & & \multicolumn{3}{c}{Fine targets} & \multicolumn{3}{c}{Coarse targets} & All \\
\cmidrule(lr){3-5}\cmidrule(lr){6-8}\cmidrule(lr){9-9}
Method & Answers & C-100 & Tiny IN & Mean & C-10/STL & C-100-20 & Mean & Mean \\
 & & (14) & (6) & (20) & (4) & (2) & (6) & (26) \\
\midrule
\multicolumn{9}{l}{\emph{No answers}} \\
Louvain base (count-free) & 0 & 62.8 & 56.2 & 60.8 & 47.6 & 32.0 & 42.4 & 56.6 \\
Name groups & 0 & 72.8 & 71.8 & 72.5 & 55.7 & 25.9 & 45.8 & 66.3 \\
LLM merge$^\dagger$ & 0 & 72.8 & 70.7 & 72.2 & 57.0 & 25.9 & 46.6 & 66.3 \\
Name groups + cleaning & 0 & 76.9 & 74.1 & 76.0 & 60.6 & 27.4 & 49.5 & 69.9 \\
\midrule
\multicolumn{9}{l}{\emph{True class count given, no answers}} \\
KMeans, base features & $K$ & 51.7 & 48.5 & 50.8 & 70.2 & 40.3 & 60.2 & 53.0 \\
Louvain, count-matched & $K$ & 59.2 & 57.3 & 58.7 & 87.1 & 37.6 & 70.6 & 61.4 \\
KMeans, CLIP-L features & $K$ & 46.6 & 46.9 & 46.7 & 67.4 & 33.2 & 56.0 & 48.9 \\
\midrule
\multicolumn{9}{l}{\emph{Ten answers, before cleaning}} \\
Contact queries & 10 & \textbf{75.8} & 72.5 & \textbf{74.8} & 83.5 & 32.5 & 66.5 & 72.9 \\
Calibrated merge queries$^\dagger$ & 10 & \textbf{75.8} & \textbf{72.6} & \textbf{74.8} & 78.1 & 31.8 & 62.6 & 72.0 \\
\textbf{LevelFirst} & 10 & 75.4 & 72.4 & 74.5 & \textbf{93.0} & \textbf{49.3} & \textbf{78.4} & \textbf{75.4} \\
\midrule
\multicolumn{9}{l}{\emph{Ten answers, after cleaning}} \\
Calibrated merge$^\dagger$ + cleaning & 10 & \textbf{78.9} & 73.9 & \textbf{77.4} & 80.5 & 33.2 & 64.7 & 74.5 \\
\textbf{LevelFirst} & 10 & 78.4 & \textbf{74.2} & 77.1 & \textbf{93.7} & \textbf{50.8} & \textbf{79.4} & \textbf{77.6} \\
\midrule
\multicolumn{9}{l}{\emph{Twenty answers, before cleaning}} \\
Contact queries & 20 & \textbf{77.2} & \textbf{73.0} & \textbf{75.9} & 89.2 & 44.4 & 74.3 & 75.5 \\
LevelFirst & 20 & 76.2 & 72.9 & 75.2 & \textbf{96.2} & \textbf{52.0} & \textbf{81.5} & \textbf{76.7} \\
\bottomrule
\end{tabular}

%% file: generated/secondary.tex
\begin{tabular}{lrrrrrrr}
\toprule
Sets & ARI & ACC & NMI & Clusters & Classes & Kept & Kept purity \\
\midrule
CIFAR-100 (14) & 78.4 & 84.3 & 86.9 & 47 & 50 & 74.1 & 94.2 \\
Tiny IN (6) & 74.2 & 81.9 & 83.5 & 89 & 100 & 71.4 & 94.5 \\
Fine (20) & 77.1 & 83.6 & 85.9 & 60 & 65 & 73.3 & 94.3 \\
Coarse (6) & 79.4 & 77.9 & 86.3 & 31 & 13 & 86.9 & 99.2 \\
All (26) & 77.6 & 82.3 & 86.0 & 53 & 53 & 76.4 & 95.4 \\
\bottomrule
\end{tabular}

%% file: generated/levels.tex
\begin{tabular}{lrrrrr}
\toprule
Dataset & Sets & Specific level & Specific names & After level queries & Change \\
\midrule
CIFAR-100 & 14 & 10 of 14 & 72.8 & 73.4 & +0.6 \\
Tiny ImageNet & 6 & 6 of 6 & 71.8 & 71.8 & +0.0 \\
CIFAR-10 & 2 & 0 of 2 & 58.0 & 74.6 & +16.6 \\
STL-10 & 2 & 0 of 2 & 53.4 & 83.1 & +29.6 \\
CIFAR-100-20 & 2 & 0 of 2 & 25.9 & 39.4 & +13.5 \\
\bottomrule
\end{tabular}

%% file: generated/noise.tex
\begin{tabular}{lrrrr}
\toprule
Ten answers, ARI before cleaning (26 sets, 5 seeds) & $p=0$ & $0.1$ & $0.2$ & $0.3$ \\
\midrule
Contact queries & 72.9 & 71.1 & 69.6 & 67.6 \\
Calibrated merge queries$^\dagger$ & 72.0 & 69.9 & 68.4 & 65.9 \\
LevelFirst & 75.4 & 73.1 & 70.9 & 68.2 \\
\quad with recheck of split level answers & 75.0 & 73.7 & 71.8 & 69.7 \\
\quad with two extra level queries & 75.0 & 73.0 & 71.0 & 69.6 \\
\midrule
Name groups, no answer & \multicolumn{4}{c}{66.3} \\
\midrule
LevelFirst after cleaning (12 fresh-comparison sets, seed 0) & 78.5 & 75.1 & 73.9 & 70.8 \\
Name groups + cleaning, no answer (12 sets) & \multicolumn{4}{c}{63.1} \\
\bottomrule
\end{tabular}

%% file: generated/criteria.tex
\begin{tabular}{llrlr}
\toprule
Criterion & Kind & Value & Bar & Outcome \\
\midrule
\multicolumn{5}{l}{\emph{Held-out evaluation of level queries (12 sets)}} \\
Before cleaning: beats ten contact queries & primary & 70.6 & $\ge$ 70.1 & passed \\
Before cleaning: no cost on the ten fine sets & primary & 74.8 & $\ge$ 74.7 & passed \\
After cleaning: beats calibrated-merge pipeline & primary & 73.2 & $\ge$ 73.1, 70.1 & passed \\
CIFAR-100-20, before cleaning: beats contact queries & secondary & 49.3 & $\ge$ 37.5 & passed \\
CIFAR-100-20: general level chosen on both sets & secondary & -- & -- & passed \\
\multicolumn{5}{l}{\emph{Fresh comparison: level + calibrated merge as a combined variant (12 sets)}} \\
Combined pipeline, fresh fine sets & primary & 76.6 & $\ge$ 76.7, 78.1 & \textbf{failed} \\
Combined pipeline, coarse sets & primary & 77.8 & $\ge$ 69.7, 78.4 & \textbf{failed} \\
Combined pipeline beats both, all 26 sets & secondary & 77.1 & $>$ 74.5, 77.6 & \textbf{failed} \\
\multicolumn{5}{l}{\emph{Noisy users (before cleaning: 26 sets, 5 seeds; after cleaning: 12 fresh-comparison sets, seed 0)}} \\
Before cleaning, $p=0.2$: beats no answers by 2 & primary & 70.9 & $\ge$ 68.3 & passed \\
Before cleaning: loses at most 2 at $p=0.1$ & primary & 73.1 & $\ge$ 73.4 & \textbf{failed} \\
After cleaning, $p=0.2$: beats no answers by 1 & primary & 73.9 & $\ge$ 64.1 & passed \\
Before cleaning: best of three strategies & secondary & -- & -- & passed \\
\multicolumn{5}{l}{\emph{Answer checking for level queries (26 sets, 5 seeds)}} \\
Escalate: loses at most 2 at $p=0.1$ & primary & 73.0 & $\ge$ 73.4 & \textbf{failed} \\
Escalate: no cost at $p=0$ & primary & 75.0 & $\ge$ 74.9 & passed \\
Escalate: gain at $p=0.2$ & secondary & 71.0 & $\ge$ 71.4 & \textbf{failed} \\
Escalate: halves flipped defaults at $p=0.1$ (\% of runs) & secondary & 11.5 & $\le$ 3.5 & \textbf{failed} \\
Recheck: loses at most 2 at $p=0.1$ & primary & 73.7 & $\ge$ 73.4 & passed \\
Recheck: no cost at $p=0$ & primary & 75.0 & $\ge$ 74.9 & passed \\
Recheck: gain at $p=0.2$ & secondary & 71.8 & $\ge$ 71.4 & passed \\
Recheck: halves flipped defaults at $p=0.1$ (\% of runs) & secondary & 9.2 & $\le$ 3.5 & \textbf{failed} \\
\bottomrule
\end{tabular}

%% file: generated/per_set.tex
\begin{tabular}{llcrrrrrrrrr}
\toprule
Set & Role & Level & Classes & Images & Louv. & Names & LLM$^\dagger$ & Cont. & LF in. & Cal.+cl.$^\dagger$ & LF \\
\midrule
C-100 s0 B & dev. & spec. & 50 & 25{,}000 & 51.3 & 67.1 & 67.6 & \textbf{68.3} & 68.0 & \textbf{72.2} & 70.9 \\
C-100 s0 LT & dev. & spec. & 50 & 5{,}499 & 63.2 & 70.2 & 70.3 & 74.9 & \textbf{75.0} & \textbf{78.2} & 77.7 \\
C-100 s1 B & held-out & spec. & 50 & 25{,}000 & 63.7 & 71.1 & 71.1 & \textbf{72.8} & 72.7 & \textbf{76.6} & 76.0 \\
C-100 s1 LT & held-out & spec. & 50 & 5{,}499 & 66.7 & 74.4 & 74.4 & \textbf{76.7} & 75.7 & \textbf{79.5} & 77.4 \\
C-100 s2 B & held-out & gen. & 50 & 25{,}000 & 56.9 & 67.7 & 67.7 & \textbf{69.3} & 68.5 & \textbf{73.8} & 71.9 \\
C-100 s2 LT & held-out & gen. & 50 & 5{,}499 & 69.2 & 78.0 & 78.0 & 80.1 & \textbf{80.3} & \textbf{86.2} & 83.4 \\
C-100 s3 B & held-out & spec. & 50 & 25{,}000 & 63.5 & 72.6 & 72.9 & \textbf{74.7} & 73.8 & \textbf{79.9} & 79.4 \\
C-100 s3 LT & held-out & spec. & 50 & 5{,}499 & 72.6 & 81.3 & 81.3 & \textbf{86.3} & 85.9 & \textbf{88.1} & 87.9 \\
C-100 s4 B & held-out & spec. & 50 & 25{,}000 & 59.5 & 70.9 & 70.9 & \textbf{71.8} & 71.3 & \textbf{76.9} & 75.9 \\
C-100 s4 LT & held-out & gen. & 50 & 5{,}499 & 65.5 & 72.5 & 72.5 & 78.2 & \textbf{78.7} & \textbf{81.2} & 80.7 \\
C-100 s5 B & fresh & spec. & 50 & 25{,}000 & 54.3 & 69.7 & 67.2 & \textbf{70.9} & 69.9 & 74.2 & \textbf{74.9} \\
C-100 s5 LT & fresh & spec. & 50 & 5{,}499 & 71.2 & 77.4 & 78.9 & \textbf{83.7} & 83.4 & 80.8 & \textbf{83.2} \\
C-100 s6 B & fresh & spec. & 50 & 25{,}000 & 52.3 & 70.3 & 69.9 & \textbf{70.6} & 69.9 & 73.7 & \textbf{74.2} \\
C-100 s6 LT & fresh & gen. & 50 & 5{,}499 & 69.2 & 75.8 & 75.8 & 82.2 & \textbf{82.7} & \textbf{83.8} & 83.4 \\
\addlinespace
Tiny IN s0 B & dev. & spec. & 100 & 50{,}000 & 48.5 & 65.9 & 64.8 & \textbf{66.3} & 66.1 & 68.2 & \textbf{69.0} \\
Tiny IN s0 LT & dev. & spec. & 100 & 10{,}847 & 65.7 & 77.5 & 77.5 & \textbf{79.1} & 78.9 & 80.8 & \textbf{81.3} \\
Tiny IN s1 B & held-out & spec. & 100 & 50{,}000 & 48.8 & 68.5 & 68.3 & \textbf{69.0} & 68.9 & \textbf{71.3} & 71.2 \\
Tiny IN s1 LT & held-out & spec. & 100 & 10{,}847 & 59.2 & 72.2 & 67.2 & \textbf{72.9} & 72.6 & \textbf{73.7} & \textbf{73.7} \\
Tiny IN s2 B & fresh & spec. & 100 & 50{,}000 & 50.6 & 68.5 & 68.5 & \textbf{69.3} & 69.0 & 69.8 & \textbf{70.2} \\
Tiny IN s2 LT & fresh & spec. & 100 & 10{,}847 & 64.7 & 78.2 & 78.2 & 78.7 & \textbf{78.9} & \textbf{79.8} & 79.5 \\
\addlinespace
C-10 B & dev. & gen. & 10 & 50{,}000 & 73.1 & 59.4 & 59.4 & 74.6 & \textbf{92.8} & 78.9 & \textbf{93.6} \\
C-10 LT & dev. & gen. & 10 & 12{,}406 & 33.7 & 56.6 & 56.8 & \textbf{95.4} & 95.3 & 93.9 & \textbf{95.3} \\
\addlinespace
STL-10 B & dev. & gen. & 10 & 13{,}000 & 44.6 & 46.8 & 51.7 & 66.8 & \textbf{85.0} & 65.6 & \textbf{87.2} \\
STL-10 LT & dev. & gen. & 10 & 3{,}223 & 39.1 & 60.1 & 60.2 & 97.2 & \textbf{98.8} & 83.5 & \textbf{98.7} \\
\addlinespace
C-100-20 B & held-out & gen. & 20 & 50{,}000 & 37.4 & 27.8 & 27.8 & 34.5 & \textbf{42.3} & 35.6 & \textbf{44.3} \\
C-100-20 LT & held-out & gen. & 20 & 11{,}514 & 26.6 & 24.0 & 24.0 & 30.4 & \textbf{56.3} & 30.7 & \textbf{57.2} \\
\bottomrule
\end{tabular}

%% file: generated/prompts.tex
\paragraph{VLM naming prompt (Qwen2.5-VL, one call per community).}
\begin{quote}\small\ttfamily
Images 1 to \textless{}k\textgreater{} each show eight small photos from one group of photos. List the distinct kinds of object shown across all photos, at most four. For each give a specific name and a general name, common nouns of one to three words. Answer with JSON only: \{"kinds": [\{"specific": "...", "general": "..."\}]\}. Example item: \{"specific": "acoustic guitar", "general": "guitar"\}.
\end{quote}

\paragraph{CLIP-L text templates (the name replaces \{\}).}
\begin{quote}\small\ttfamily
itap of a \{\}.\\
a bad photo of the \{\}.\\
a origami \{\}.\\
a photo of the large \{\}.\\
a \{\} in a video game.\\
art of the \{\}.\\
a photo of the small \{\}.
\end{quote}

\paragraph{Judge ladder of the calibrated-merge baseline (gpt-oss).}
\begin{quote}\small\ttfamily
Two groups of photos are labelled '\textless{}name A\textgreater{}' and '\textless{}name B\textgreater{}'. \textless{}question\textgreater{} Answer with JSON only: \{"same": true or false\}.\\
L1: Do both names denote the same specific variety, breed or model?\\
L2: Do these names describe the same kind of object, so that a person sorting photos by kind of object would put both groups together?\\
L3: Do both names belong to the same broad category of things?
\end{quote}

\paragraph{LLM merge baseline: caption prompt (VLM).}
\begin{quote}\small\ttfamily
Guideline: Group the photos by the kind of object they mainly show. Describe the main object in this photo in one short sentence, focusing on what kind of object it is.
\end{quote}

\paragraph{LLM merge baseline: merge prompt (gpt-oss).}
\begin{quote}\small\ttfamily
Guideline: Group the photos by the kind of object they mainly show.\\
Cluster A photos:\\
\textless{}captions of cluster A\textgreater{}\\
Cluster B photos:\\
\textless{}captions of cluster B\textgreater{}\\
Under the guideline, should cluster A and cluster B be merged into one group? Answer with JSON only: \{"merge": true or false\}.
\end{quote}